\documentclass[sn-nature]{sn-jnl}

\usepackage{graphicx}%
\usepackage{multirow}%
\usepackage{amsmath,amssymb,amsfonts}%
\usepackage{amsthm}%
\usepackage{mathrsfs}%
\usepackage[title]{appendix}%
\usepackage{xcolor}%
\usepackage{textcomp}%
\usepackage{manyfoot}%
\usepackage{booktabs}%
\usepackage{algorithm}%
\usepackage{algorithmicx}%
\usepackage{algpseudocode}%
\usepackage{listings}%

\theoremstyle{thmstyleone}%
\theoremstyle{thmstyletwo}%

\theoremstyle{thmstylethree}%

\begin{document}

\title[Article Title]{Personalization of Stress Mobile Sensing using
Self-Supervised Learning}


\author*[1]{\fnm{Tanvir} \sur{Islam}}\email{tislam@hawaii.edu}

\author*[1]{\fnm{Peter} \sur{Washington}}\email{pyw@hawaii.edu}

\affil*[1]{\orgdiv{Information and Computer
Sciences Department}, \orgname{University of Hawai’i at Manoa}, \orgaddress{\street{1680 East-West Road}, \city{Honolulu}, \postcode{96822}, \state{Hawaii}, \country{USA}}}


\abstract{Stress is widely recognized as a major contributor to a variety of health issues. Stress prediction using biosignal data recorded by wearables is a key area of study in mobile sensing research because real-time stress prediction can enable digital interventions to immediately react at the onset of stress, helping to avoid many psychological and physiological symptoms such as heart rhythm irregularities. Electrodermal activity (EDA) is often used to measure stress. However, major challenges with the prediction of stress using machine learning include the subjectivity and sparseness of the labels, a large feature space, relatively few labels, and a complex nonlinear and subjective relationship between the features and outcomes. To tackle these issues, we examine the use of model personalization: training a separate stress prediction model for each user. To allow the neural network to learn the temporal dynamics of each individual's baseline biosignal patterns, thus enabling personalization with very few labels, we pre-train a 1-dimensional convolutional neural network (CNN) using self-supervised learning (SSL). We evaluate our method using the Wearable Stress and Affect prediction (WESAD) dataset. We fine-tune the pre-trained networks to the stress prediction task and compare against equivalent models without any self-supervised pre-training.  We discover that embeddings learned using our pre-training method outperform supervised baselines with significantly fewer labeled data points: the models trained with SSL require less than 30\% of the labels to reach equivalent performance without personalized SSL. This personalized learning method can enable precision health systems which are tailored to each subject and require few annotations by the end user, thus allowing for the mobile sensing of increasingly complex, heterogeneous, and subjective outcomes such as stress.}




\maketitle

\section{Introduction}
Chronic stress can drastically affect an individual's health across several dimensions. Continuous stress increases the risk of cardiovascular disease, hypertension, reduced immunity, and cancer \cite{cohen2007psychological}. Despite these detrimental physical effects, stress is often left unmanaged. A growing body of work in the field of human-computer interaction (HCI) is focusing on mobile sensing of stress using signals from wearable consumer devices \cite{paredes2011calmmenow, sarsenbayeva2019measuring, hansel2018put, wang2020social, liapis2015recognizing, zakaria2019stressmon}. Real-time detection of stress via wearable devices can enable stress interventions which can provide therapeutic support via immediate prompting when stress is detected. Until recently, these mobile sensing methods of subjective outcomes have required large amounts of training data from each user, making the deployment of such models infeasible in practice. We present a novel methodology for personalizing mobile sensing models using a minimal number of human labels. 

Consumer wearable devices offer a range of biosignal measurements, which may include but are not limited to EDA, electrocardiograms (ECG), electromyography (EMG), respiration rate, and galvanic skin resistance (GSR). EDA is a physiological measurement of the electric current that flows through the skin. EDA is particularly sensitive to changes in skin ionic conductivity caused by even minor perspiration that is not visible on the surface of the skin \cite{giannakakis2019review}. In contrast to other organs in the body that are connected to the parasympathetic branch, human skin, including its sweat glands and blood vessels, is exclusively supplied by the sympathetic nervous system \cite{boucsein2012electrodermal}. In comparison to other physiological measurements like heart rate variations and blood pressure, EDA is a robust way to gauge physiological arousal and, by extension, the response to stress. In the field of affective computing, EDA is widely recognized as a key marker of physiological arousal and stress responses \cite{1438384Healey}\cite{articleSetz}. Therefore, an EDA-based stress quantification approach is the primary focus of this investigation.

Deep neural networks (DNNs) have been a driving force behind major machine learning progress in recent years \cite{lecun2015deep}. DNNs are multi-layer computational models that may learn progressively complex and high-level representations of the input data to accomplish a prediction goal. DNNs have been revolutionary in fields as diverse as speech recognition \cite{pmlr-v48-amodei16}, computer vision \cite{7410480}, NLP, strategic games \cite{silverD}, and medical domain \cite{10.1001/jama.2016.17216}. DNNs are an especially promising solution for analyzing biosignals due to their ability to spot patterns and learn meaningful features from unprocessed data inputs without requiring extensive data preprocessing or manual feature engineering.

Creating a stress prediction model using biosignal data that is generalizable is impractical due to the varying physiological responses to stress among individuals. As a result, stress monitoring systems must possess the flexibility to accommodate these differences \cite{schmidt2018wearable}. Rather than building a one-size-fits-all stress prediction model, we propose a novel approach aimed at developing individual-specific models, with the goal of achieving greater personalization. Such models can enable precision health mobile sensing systems to reach clinically useful performance.

A major hurdle in developing personalized mobile sensing models through biosignal data is the procurement of adequate supervisory data (i.e., ground truth labels) to train DNNs, which require large training sets for the training process to converge. Self-reported stress labels are often generated by laborious and faulty manual human labeling owing to human subjectivity and weariness \cite{10.1162/qss_a_00144}. 

Current research on EDA-based stress assessment tends to focus on supervised deep learning. Achieving clinically useful model performance is hindered, however, by the fact that traditional supervised learning does not make use of all of the unlabeled data that are available. Individuals who use wearable devices have copious amounts of unlabeled longitudinal body signal data, but only a few labels to indicate important health events. These labels are used for training machine learning models to recognize the events of interest. 

To bridge this gap, we propose the use of SSL to learn neural network weights which are optimized to make predictions about the baseline temporal dynamics of a user's biosignals. Such weights can then be efficiently fine-tuned to personalized prediction tasks using that user's biosignal data. We implement an instantiation of this SSL paradigm and then fine-tune the resulting models to the stress prediction task using EDA signals. In particular, we make the following contributions:

\begin{itemize}
    \item We introduce a novel ``personalized SSL'' paradigm to address the common real-world situation of large datasets with few training labels. This solution enables mobile sensing systems to predict traditionally subjective classes such as stress. This interaction paradigm is similar to how Apple's Face ID works: once the face recognition system is provided with only a few examples provided by the end users, it can be used repeatedly to unlock the device. In this case, the labels are stress predictions rather than identity predictions and the model input is wearable biosignals rather than a face image. 
    \item We evaluate this novel mobile sensing methodology on a real-world biosignals dataset with stress labels: the Wearable Stress and Affect Detection (WESAD).

Once the model has been calibrated, it can be used repeatedly to predict how stressed that person will be in different scenarios. The stress prediction task requires less computing power because the model uses the model weights learned during the training process to efficiently fine-tune towards stress prediction. This reduced need for computing resources means that the model can be processed faster, making it easier to use in real-world settings.

\end{itemize}

\section{Related Work}

\subsection{Stress Assessment using Biosignals}

Hyperactivity of the nervous system is associated with chronic stress and is responsible for its wide range of adverse effects on health and behavior. Neuroimaging research has shown a possible connection between HRV and areas of the brain that are responsible for assessing stressful situations such as the ventromedial prefrontal cortex \cite{kim2018stress}. It was hypothesised by He et al. that HRV-based approaches may be used to identify acute cognitive stress, who demonstrated that acute cognitive stress may be detected in real-time using HRV \cite{he2019real}. 

Stress and brain activity are closely correlated \cite{Dharmawan2007AnalysisOC}. Single-channel electroencephalography (EEG) headsets, mental strain, and a public-speaking assignment serve as stimuli in research which attempts to quantify the range of mental stress experienced by study participants \cite{secerbegovic2017mental}. Statistical analysis of EEG data reveals the existence of the alpha and beta bands which are useful for classifying stress. To quantify mental strain, researchers have suggested using EEG and the Hilbert-Huang transform as input features for a support vector machine (SVM) classifier \cite{vanitha}. It has been demonstrated by Halim et al. that there exists a framework for measuring driver stress using EEG \cite{HALIM202066}. EEG data were captured to monitor the driver's continuing brainwave activity while driving with the goal of establishing a relationship between the driver's brain activity and their emotional reaction \cite{HALIM202066}.

EMG is a biological signal that measures the electrical charge generated by engaging human muscles to reflect the neuromuscular activity. Karthikeyan et al. discuss the results of an EMG investigation that analyzed the link between a person's fluctuating stress levels and the tension in their muscles \cite{karthikeyan2012emg}.

ECG is the most common biosignal used for monitoring heart activity. ECG is not only a reliable indicator of heart rate but also of stress levels in humans \cite{ahn2019novel}. A method for gauging human stress using a mono-fuzzy index derived from ECG data is described in \cite{8512499}. In \cite{walocha2022activity}, Walocha et al. looked at ECG data and presented a person-independent and a person-specific strategy to stress prediction, respectively. The authors studied subjects to stress while driving manually while multitasking, and while driving autonomously. Hannun et al. show that single-lead ECGs can be used to identify rhythms using an end-to-end deep learning method. They demonstrate that the 1D CNN model architecture can be successfully applied towards making predictions from biosignal data without manual feature engineering and feature selection \cite{hannun2019cardiologist}.

Variations in the skin's electrical characteristics are reflected by EDA, making it a helpful stress indicator for the nervous system. Sweating is a natural response to stressful situations and may further enhance the skin's conductivity \cite{wickramasuriya2018state}. Spathis et al. describe a stress prediction framework that enables consumers to receive feedback on their stress levels through speech-enabled consumer devices like smartphones and smart speakers \cite{spathis2021self}. Using data from the Empatica E4 device included in the publicly available WESAD dataset \cite{schmidt2018introducing}, Spathis et al. offer a stress prediction framework triggered by issues such as money, work, and personal responsibility. The design of a wearable sensor system for mental stress prediction is presented in \cite{affanni2020wireless}, which includes a two-channel EDA sensor and a two-channel ECG sensor. EDA sensors are located on the hands, whereas ECG sensors are located on the chest. Data from the EDA sensors is wirelessly transferred to the ECG sensor through WiFi. This method attempts to improve stress prediction by capturing two EDA channels to eliminate motion artifacts. The designed sensor yields excellent linearity and jitter performance and exhibits minimal delays during wireless transmission. Zhu et al. studied the possibility of detecting stress using EDA data acquired from wearable sensors\cite{zhu2022feasibility}. The study makes use of two publicly accessible datasets: VerBIO \cite{articleverbio} and WESAD \cite{schmidt2018introducing}, both of which comprise of EDA signals from research-grade wearable sensors. The experimental results demonstrate that Random Forests differentiate stress from non-stress conditions with an accuracy of 85.7\%, highlighting the potential for wearable devices with EDA sensors to anticipate stress. The study also evaluates the dependability of training with extracted statistical and EDA-related features from EDA signals, discovering no consistent trend between employing complete features and chosen features. The findings indicate the effectiveness of placing EDA sensors into wearable devices for stress prediction, instilling confidence that incorporating EDA sensors into smartwatches is a potential option for future product development. Sanchez-Reoli et al. review stress prediction systems based on EDA and machine learning (ML). Sanchez-Reoli et al. present two basic ways for evaluating an individual's emotional state after reviewing 58 chosen articles: utilizing only EDA for prediction and integrating EDA with additional physiological data \cite{sanchez2020machine}. Both approaches have pros and cons, with the former providing the advantages of compact non-invasive equipment and high precision, while the latter provides an improved mapping of a subject's physical, psychological, and cognitive condition.

In addition to research grade equipment, EDA is also frequently measured using consumer wearable devices. Using low-resolution Electrodermal Activity (EDA) data from consumer-grade wearable devices, Ninh et al. explore the stress prediction capabilities of user-dependent and user-independent models \cite{ninh2022analysing}. This work reveals that stress prediction models employing low-resolution EDA data from wrist-worn and finger-mounted sensors have high balanced accuracy scores (66.10\% to 100\%). This finding indicates that low-resolution EDA signals from consumer-grade devices may be used to efficiently create user-dependent stress prediction models, enabling the collection of in-the-wild data for mental health tracking and analysis. Anusha et al. show that EDA collected from a wrist device can be used to automatically recognize stress before surgery \cite{anusha2019electrodermal}. The strategy uses a supervised machine learning method to find motion glitches in the EDA data and a localized supervised learning model to predict stress. Anusha et al. demonstrate a novel method for the use of physiological changes to measure stress in a hospital setting, achieving a high level of accuracy in detecting stress.

EDA is often used in conjunction with other biosignals. Eren et al. suggest a way to detect stress by using sensors from the Empatica E4 device used in the WESAD \cite{schmidt2018introducing} dataset to measure blood volume pulse (BVP) and electrodermal activity (EDA) \cite{eren2022stress}. The suggested neural network model is 96.26\% accurate. The study also calls for more research to test the method's usefulness and general flexibility. Pakarinen et al. \cite{8857621Pakarinen} explored the capacity of electrodermal activity (EDA) to precisely categorize mental states, including relaxation, arousal, and stress, in addition to self-perceived stress and arousal, utilizing a three-phase modified MIST test. The research demonstrated high levels of accuracy in identifying these mental states, indicating that EDA has the potential for long-term evaluation of stress and arousal in the workplace. They explored the utilization of machine learning models for stress detection and classification, employing features derived from physiological signals, including EDA and BVP. By achieving an F1 score of 0.71, the study highlights the potential to create a seamless, real-world stress monitoring system and promotes continued investigation through the integration of machine learning and signal processing methodologies in a professional context. Kalimeri et al. propose an advanced multimodal framework to assess the emotional and cognitive experiences of visually impaired individuals while navigating unknown indoor environments, leveraging mobile monitoring and the combination of EEG and EDA signals \cite{kalimeri2016exploring}. The goal is to identify the environmental factors that increase stress and cognitive load, thereby guiding the creation of emotionally intelligent mobility systems that can adapt to challenging environments using real-time biosensor information.

The study of stress prediction using biosignals is a growing area of interest in the HCI community. Lee et al. created a wearable in- and over-ear gadget that monitors EEG and ECG signals concurrently \cite{lee2020stress}. Convolutional neural network (CNN) was utilized to distinguish "stressed" and "relaxed" based on EEG and ECG data with an accuracy of around 75\%. Hernandez et al. devised a continuous monitoring method for stress \cite{hernandez2014under}. In this research, two types of stress have been examined: cognitive load and subjective and personal stress. This study involved the use of a pressure-sensitive keyboard and a capacitive mouse. They concluded that in stressed conditions, typing pressure increases dramatically and there is greater contact with the mouse's surface. Sarker et al. introduced a time-series pattern matching approach to identify major stress occurrences in a time series with discontinuous and quickly variable stress data \cite{sarker2016finding}. This is the first method for analyzing stress data in the time-series domain, which helps determine the timing of just-in-time stress interventions, which are the ultimate goal of stress mobile sensing systems.

\subsection{Mobile Sensing Systems}

There are several mobile sensing systems in the HCI literature for detecting psychiatric and mental health outcomes using mobile and wearable data streams. FEEL is a personalized mobile phone application developed by Ayzenberg et al. which tracks the user's social interactions and biological data \cite{ayzenberg2012feel}. They develop customized stress identification software using real-time data and assign anxiety levels to every occurrence. Both environmental data and stress levels are pooled in an accessible notebook in which the user may represent his/her physiological reactions. This study is focused on stress vs. non-stress conditions, and the evaluation was done by a single user. More subjects would do a better job of evaluating personalization.

Hinckley et al. discussed sensing strategies inspired by special elements of HCI with mobile devices \cite{hinckley2000sensing}. Hinckley et al. conducted experiments on various mobile device interactions, such as recording memos, switching between portrait and landscape orientation, and auto-powering the device, and collected data on how people responded to these functions. It is hard to personalize as these things move forward to generalization.

MoodScope is a smartphone application that determines the user's mood based on their use behavior \cite{likamwa2013moodscope}. LiKamWa et al. demonstrate that a personalized model is a sole alternative for determining a user's emotional state, but a generic model fails when personalization is the essential component. By learning about its user and linking smartphone use patterns with certain emotions, MoodScope seeks to utilize trends such as where the user has been, with whom they speak, and what apps they use. However, it is not always possible to sense the emotion with the type of using a phone. There is a possibility to fake this kind of data. In those cases, biosensors, directly connected to the human body help as it is very hard to fake.

Experience sampling techniques (ESM) are a common technique for collecting emotional state data from users. Kang et al. conducted the first research on the consequences of ESM-introduced interrupts \cite{kang2022understanding}, demonstrating that 38\% of participants' emotional state was influenced by the ESM surveys while they were completing them \cite{kang2022understanding}. The challenge of ESM is that the records can be biased due to fluctuating emotional states.

\begin{figure*}
  
  \includegraphics[width=\linewidth]{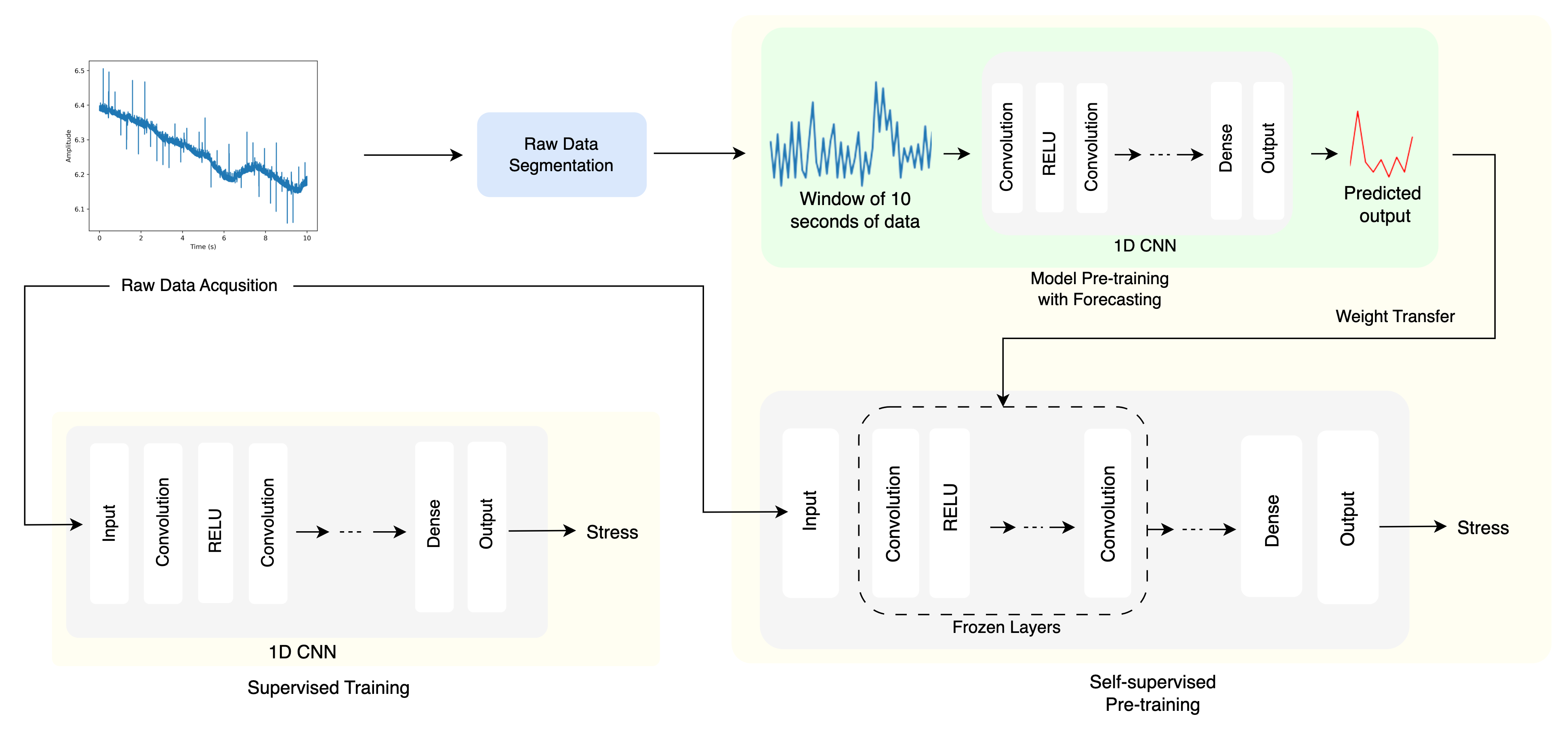}
  \caption{Overall process of our personalized stress prediction framework. The raw signal is segmented into distinct but overlapping data points. A forecasting model is used for self-supervised pre-training. These weights are fine-tuned to the stress prediction task (right model). We compare this model against a supervised learning model without self-supervised pre-training (left).  }
\label{architecture}
\end{figure*}

\subsection{SSL on Biosignals}

The field of SSL on biosignals is an emerging field of study. Two popular SSL approaches, SimCLR \cite{chen2020simple} and BOYL \cite{grill2020bootstrap}, are contrastive learning approaches which rely on data augmentation. This methodology contains limitations such as invariance, which can obscure valuable detailed information necessary for specific downstream tasks. Additionally, predicting which augmentations will yield optimal benefits can prove to be challenging \cite{zhang2022rethinking}. As a result, in our study, we have excluded the use of data augmentation and chosen instead to focus on pre-training and downstream tasks with raw data. Robert et al. offer a contrastive learning-based stress prediction system that uses modified EDA signals to classify stress and non-stress circumstances \cite{matton2022contrastive}. Nevertheless, the current project differs from this approach because it does not require any signal transformation. Instead of only stress vs. non-stress categorization, our presented study allows for the learning of a wide range of stress levels. The Step2Heart system is a self-supervised multimodal wearable data feature extractor \cite{spathis2021self}. Using the representations of heart rate in activity data, this model makes predictions about individual health outcomes using transfer learning. With the limitations of the biosignals dataset, such as the presence of noisy labels and small sample size, Cheng et al. turned to a self-supervised based on contrastive learning \cite{cheng2020subject}. By using the temporal and spatial information included within ECG impulses, CLOCS uses contrastive learning to
acquire unique representations of each patient \cite{kiyasseh2021clocs}. This novel concept is to consider a positive pair as a representation of two changed occurrences that both correspond to an identical patient. As a result, the model automatically adjusts the learnt representations to the specifics of every patient's case. Sarkar et al. offer a self-supervised method for learning ECG representations for the purpose of emotion classification \cite{sarkar2020self}. Banville et al. attempted to learn a patient representation from EHR data and proposed employing a stacked denoising autoencoder called DeepPatient \cite{banville2019self}. In this paper, Banville et al. reported the first extensive evaluation of self-supervised representation learning using 12-lead electrocardiogram data in clinical settings. To accomplish this objective, the researchers utilize state-of-the-art self-supervised algorithms that rely on instance discrimination and latent forecasting in the field of electrocardiogram (ECG) analysis. Banville et al. proposed a novel self-supervised for extracting comprehensive EEG feature sets \cite{banville2019self}. The findings of this research \cite{banville2019self} demonstrate that, particularly under limited data conditions, incorporating two key features, namely "relative positioning" and "temporal shuffling," can outperform conventional unsupervised and supervised techniques in a downstream sleep pacing task.

\section{Methodology}

We propose a two-step process for training mobile sensing models which can predict subjective labels such as stress using only a few training examples per user: (1) personalized SSL for representation learning of temporal EDA dynamics per user, and (2) fine-tuning the pre-trained model to the stress prediction task. During SSL, the pretext task helps the model learn interim data representations without ground truth labels. The pretext task is designed to predict the missing segments of the input data given the remaining segments. After this representation is learned, the downstream task involves fine-tuning the weights from the pretext task towards training the neural network model to predict stress. An overall architecture is shown in Figure~\ref{architecture}.

As depicted in Figure~\ref{architecture}, the personalized SSL model is designed to accept the raw EDA signal for each subject. The raw EDA signal is divided into chunks of 7000-dimensional vectors, as described in Section~\ref{Self-supervised Pre-training}, generating a large dataset of 7910 data points to pre-train the model. The pre-training phase involves the use of a 1D CNN model to predict 40 data points, as detailed in Section~\ref{Self-supervised Pre-training}. The weights learned in the pre-training phase are transferred to another 1D CNN model which is fine-tuned for the stress prediction task.

In addition, a second 1D CNN model with the same architecture as the fine-tuned model is employed for stress prediction, but without any pre-training. This is our control model which we use as a baseline to quantify the benefits of our SSL over personalization without SSL.

We compare these two models using varying numbers of labeled data points, as described in Section~\ref{Evaluation using pre-trained models} and Section~\ref{Results for a demonstrative user}

We perform a subject-dependent study by directly comparing the following two models trained and evaluated separately per subject:

\begin{itemize}
    \item \textbf{Model A}: Stress prediction via supervised learning
    \item \textbf{Model B}: Stress prediction with a model pre-trained with SSL and fine-tuned using the same supervised learning strategy as Model A
\end{itemize}

\subsection{Dataset}\label{dataset}
We use the WESAD \cite{schmidt2018introducing} dataset to evaluate our methodology.  WESAD \cite{schmidt2018introducing} is a freely accessible multimodal physiological dataset containing ECG, EDA, EMG, respiration (RESP), core body temperature (TEMP), and three-axis acceleration  (ACC). Biosignals were collected from 15 healthy adults (mean age 27.5; standard deviation (SD) = 2.4)) when they were subjected to one of three conditions: neutral, stress, or amusement in a controlled laboratory setting. The researchers used the Trier Social Stress Test (TSST) to induce anxiety. Two devices: Empatica E4 and RspiBAN, were used to collect signals for this dataset.  Using the RespiBAN device, data was acquired from each participant for around 33 minutes at a sample rate of 700 Hz. Multiple sensors, including an ECG, EDA, EMG, RESP, TEMP, and ACC are included in the RespiBAN device. The Empatica E4 records sensor data on BVP at 64 Hz, EDA at 4 Hz, TEMP at 4 Hz, and ACC at 32 Hz. Participants were asked six questions from the State-Trait Anxiety Inventory (STAI) to derive insight into their degree of anxiety. Each question was answered using a four-point Likert scale. The STAI questions are as follows:

\begin{itemize}
    \item Question 1: I feel at ease 
    \item Question 2: I feel nervous
    \item Question 3: I am jittery
    \item Question 4: I am relaxed
    \item Question 5: I am worried
    \item Question 6: I feel pleasant

\end{itemize}

In this study, we used stress data for baseline condition only. Our models were trained using EDA signal data from the RespiBAN device with a sampling rate of 700 HZ. 

\subsection{Label Representation}

We convert the categorical ordinal outcome variables (i.e., the Likert scale responses to the questionnaires) into a representation which maintains the categorical nature of the responses as well as the ordinal nature. In the WESAD dataset, participants were given six questions from STAI (see Section~\ref{dataset} for details). Participants were asked to rate this question on a four-point Likert scale. Therefore, each question was labelled between $1$ to $4$. To model the ordinal regression task, we convert the original labels of (1, 2, 3, 4) into the probabilities of (0.25, 0.5, 0.75, 1.0). Because the four-point Likert chart used in the original evaluation was equally spaced, these labels are quantified and distributed uniformly. These labels are subjective since models use only within-subject data and participants are internally consistent in their labeling (i.e., we assume that a rating of 3 consistently indicates higher severity than a rating of 2 on the same question within each participant's set of ratings).  

\begin{figure}[h]
  \centering
  \includegraphics[width=\linewidth]{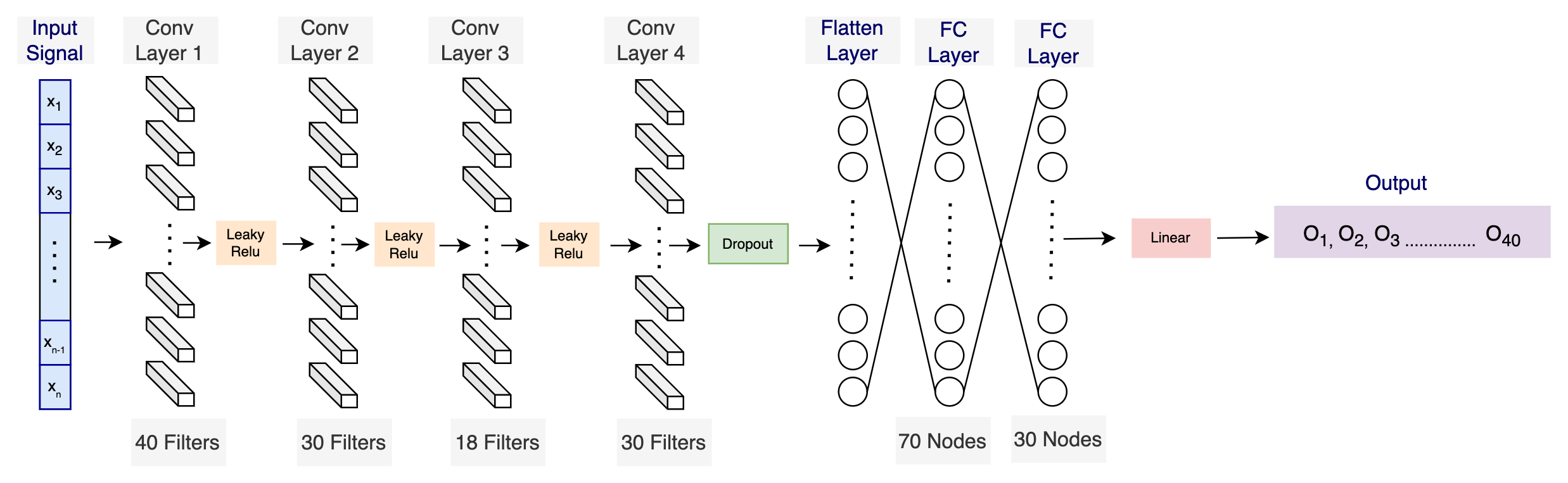}
  \caption{1D CNN architecture for pretext task. }
  \label{pretext}
\end{figure}

\subsection{Self-supervised Pre-training}\label{Self-supervised Pre-training}

The use of self-supervised pre-training, an unsupervised approach in which data is self-labeled and subsequently fine-tuned using supervised learning approaches to extract important feature representations, has proven to be more effective than supervised training alone in an increasing number of applications \cite{chen2020simple}. The self-supervised pre-training approach focuses on learning a high-level representation of data without using ground truth labels which is also known as the pretext task. A pre-trained model enables efficient transfer to a target task since the representation it has learned is already suited (or readily transferable) for a given job. SSL in this manner is the same technology which powers large-scale language models like ChatGPT and natural language training tasks powered by BERT \cite{devlin2018bert, lee2022coauthor, karinshak2023working}.

We pre-train our model through forecasting, a similar technique used to pre-train OpenAI's GPT language model \cite{radford2018improving}. A separate model was pre-trained for each human subject using only the data from that subject. To pre-train our model, we have employed a 1D CNN, a standard neural network architecture for signal prediction \cite{ancillon2022machine, kiranyaz20191, rim2020deep}. Prior research has demonstrated the robustness of 1D CNN in 1D signal processing tasks, particularly in the domain of biosignals or 1D signals \cite{kim2020hyperparameter, hannun2019cardiologist, tang2020rethinking}. One of the key advantages of 1D CNNs is their ability to automatically learn relevant features from the input signal, eliminating the need for manual feature engineering. This can save time and effort while also improving the model's performance, as it can adaptively learn the most discriminative features for a given task.

Moreover, 1D CNNs can capture information at different time scales due to the use of multiple neural layers and multiple independent learned kernels. This makes this family of architectures particularly helpful when dealing with signals with complicated hierarchical patterns or events that occur at different time scales. Additionally, 1D CNNs can handle input data with noise better than other machine learning algorithms, as the convolutional layers can learn to focus on the most important features and ignore noise or irrelevant signals. Figure~\ref{pretext} illustrates the architecture that has been used in this study for pre-training the model as a knowledge representation of the raw EDA signal. To maintain consistency, we have used the same model architecture for both self-supervised pre-training and supervised training. We use Root Mean Squared Error (RMSE) as our evaluation metric.

The pre-text 1D CNN model is implemented with an input shape of (7000, 1), consisting of four 1D convolution layers, each utilizing a leaky ReLU activation function. The first convolution layer has 40 filters, followed by a layer with 30 filters, and then a layer with 18 filters. The fourth convolution layer has 30 filters and uses a leaky ReLU activation function. Moreover, the model included two dense layers, one with 70 nodes and the other with 30 nodes. The final layer is an output layer with a linear activation function. The convolution layers are responsible for extracting important features from the EDA signal, while the dense layers utilize these features to make a stress prediction.

\begin{figure}[h]
  
  \includegraphics[width=\linewidth]{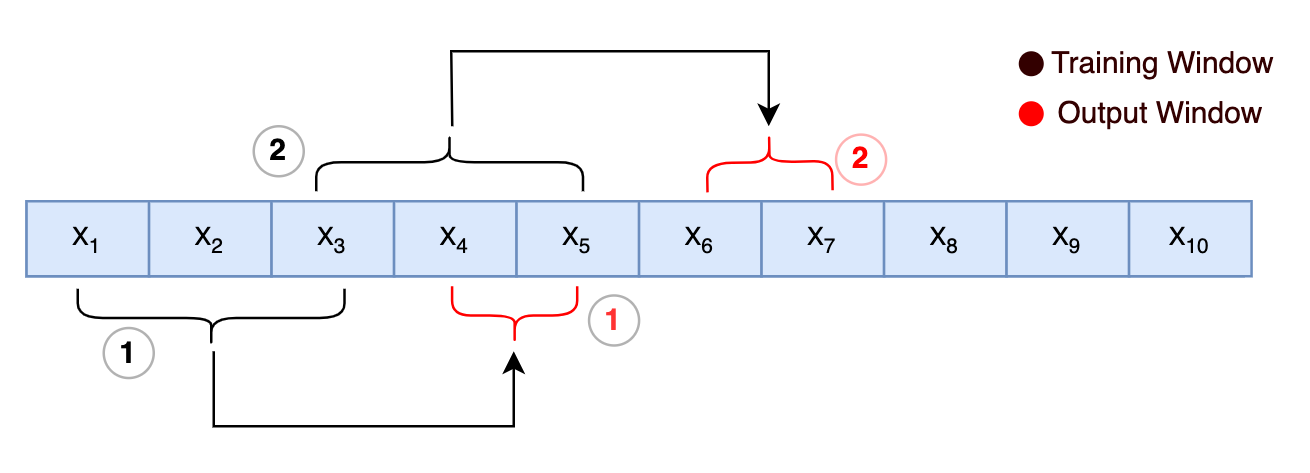}
  \caption{ We use signal forecasting as our self-supervised pretext task. A sliding window is used to predict the following chunk of the EDA signal. }
  \label{sliding}
\end{figure}

The model is pre-trained on the entirety of the EDA signal for each participant. We divide the whole signal into fixed windows (each containing a 7000-dimensional vector) (Figure~\ref{sliding}). Figure~\ref{sliding} illustrates an example of data points in the training set with overlapping information. Assuming that the signal is represented by $X_1 \cdots \cdots X_n$, each training data point includes the sequence of $X_1 \cdots \cdots X_{7000}$, and the corresponding target is $X_{7001} \cdots \cdots X_{7040}$. The training data points overlap by 100 indices, meaning that the next training data point will be $X_{100} \cdots \cdots X_{7100}$  while the corresponding target will be $X_{7101} \cdots \cdots X_{7140}$. Using this process, we produce a total of 7,910 data points for pre-training per human subject. We predict a 40-point frame using a 10-second window of the original data. The rationale of this approach is to enable the model to predict stress levels using 10-second time segments.

\subsection{Downstream Stress Prediction Task} \label{ssl}

Self-supervised pre-training offers the benefit of optimizing data efficiency during the fine-tuning process for subsequent downstream activities. The difficulty in sourcing high-quality labels makes this a critical scenario for use in mobile sensing research. We conduct a thorough quantitative analysis of this added benefit by contrasting the efficiency of freshly trained, solely supervised models against that of those pre-trained using self-supervised representations. 

\begin{figure}[h]
  
  \includegraphics[width=\linewidth]{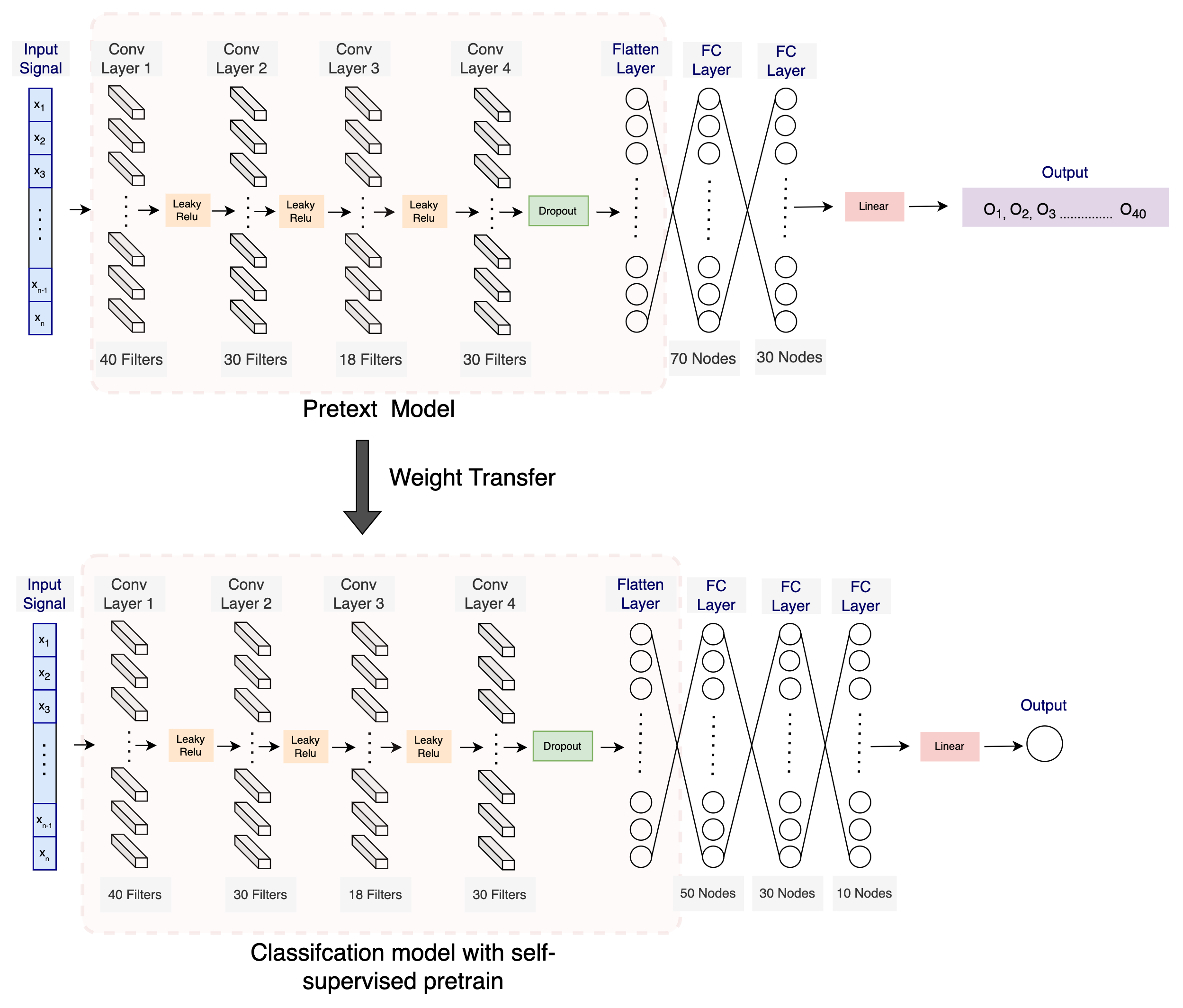}
  \caption{1D CNN architecture for fine-tuning the pre-trained network towards the downstream stress prediction task. }
  \label{classificationpretrainmodel}
\end{figure}

In this stage, the feature extractor layers are frozen, and the rest of the layers are fine-tuned to the stress prediction task. For this downstream task, we have modified the model with $3$ new dense layers for the final prediction. The architecture for stress quantifying is shown in Figure~\ref{classificationpretrainmodel}. As illustrated in Figure~\ref{classificationpretrainmodel}, the fine-tuned model shares the same set of convolution layers and activation functions as the pre-text 1D CNN model. However, it has three dense layers, the first one with $50$ nodes, followed by a dense layer with $30$ nodes, and the final dense layer with $10$ nodes. The output layer has a linear activation function, and the predicted output is stress.

In the case of purely supervised training, the same architecture as the fine-tuned model is utilized to facilitate direct comparison between methods.

\begin{table}

\caption{RMSE for the signal forecasting pre-training task for all subjects. The low RMSE indicates that the model learned the baseline dynamics of each user's EDA biosignals.} 

\begin{tabular}{|p{0.4cm}|p{0.4cm}|p{0.4cm}|p{0.4cm}|p{0.4cm}|p{0.4cm}|p{0.4cm}|p{0.4cm}|p{0.4cm}|p{0.4cm}|p{0.4cm}|p{0.4cm}|p{0.4cm}|p{0.4cm}|p{0.4cm}|}
\hline
\tiny S2 & \tiny S3 & \tiny S4 & \tiny S5 & \tiny S6 & \tiny S7 & \tiny S8 & \tiny S9 & \tiny S10 & \tiny S11 & \tiny S13 & \tiny S14 & \tiny S15 & \tiny S16 & \tiny S17 \\
\hline
\tiny 0.024 & \tiny 0.045 & \tiny 0.018 & \tiny 0.027 & \tiny 0.027 & \tiny 0.050 & \tiny 0.032 & \tiny 0.113 & \tiny 0.017 & \tiny 0.031 & \tiny 0.015 & \tiny 0.019 & \tiny 0.026 & \tiny 0.021 & \tiny 0.030 \\
\hline
\end{tabular}
\label{pretrainmodeltable}
\end{table}

\subsection{Experimental Procedures}

For our experiment, we select a fixed number of random data points having window size of $7,000$ which corresponds to $10$ seconds of data to train our model in both SSL and solely supervised manners with exactly the same set of data points. In both cases, we utilize the same model architecture and the same sampled data points. To train using SSL, we transferred the weights from the pre-trained model to the downstream task (Section \ref{ssl}). The average RMSE of the trained model using 10 data points is shown in Figure~\ref{15pointsresult}. In this experiment, all models are subjected to an identical test set.


\section{Results }

We use RMSE to evaluate our models, as regression models commonly utilize RMSE as a statistic for assessing model performance. For a given test set, a model's RMSE measures the average distance between the expected values and the actual values. It is measured as the square root of the average of the squared differences between the predicted and actual values. RMSE is helpful because it penalizes large mistakes over smaller ones and quantifies far apart the mistakes in the original units of the dataset. A smaller RMSE indicates that the model performed with less error.

\subsection{Pre-trained Models}
For pre-training purposes, all the models have been trained using data coming from a single subject. Table~\ref{pretrainmodeltable} demonstrates the RMSE of the pre-trained models of all $15$ subjects. The RMSE score is a measure of how well a model performed in predicting values compared to the actual values. For example, Subject 2's pre-trained model achieves an RMSE score of $0.024$, meaning that the model's predictions are, on average, $0.024$ units away from the actual values. In other words, the model's predictions have an average error of $0.024$ units.  The lower the RMSE score, the better the model's performance. A smaller RMSE value indicates that the model is making more accurate predictions, while a higher RMSE value indicates that the model is making larger errors. The low RMSE values we achieved across subjects suggest that our pre-trained models are well-trained and have learned the underlying dynamics of each subject's EDA signals.

\subsection{Evaluation using pre-trained models} \label{Evaluation using pre-trained models}

\begin{figure*}
  
  \includegraphics[width=\linewidth]{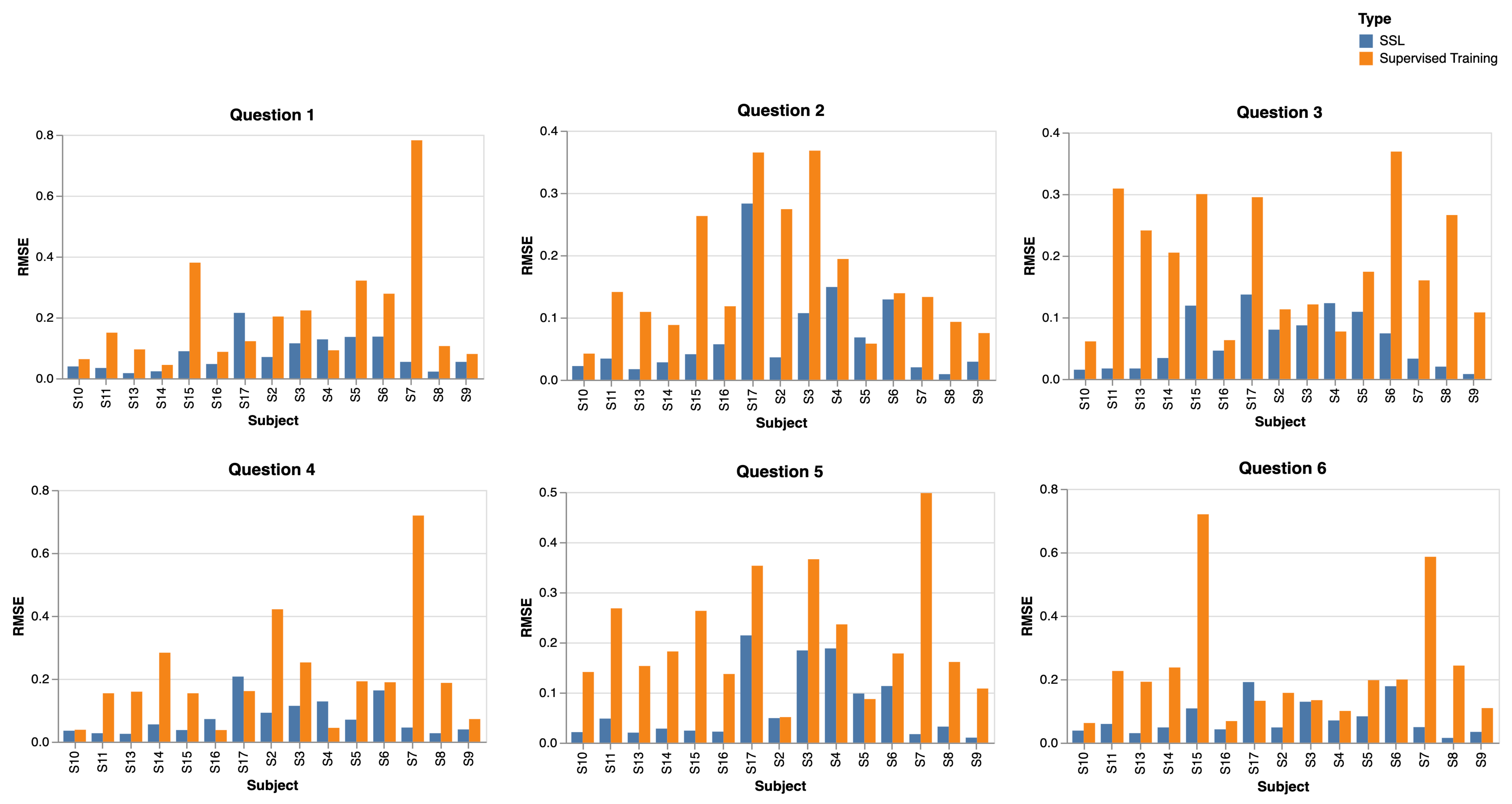}
  \caption{Comparison of the RMSE between a model fine-tuned after SSL vs. purely supervised training for all subjects and for all questions.}
  \label{15pointsresult}
\end{figure*}

Figure~\ref{15pointsresult} displays the RMSE score of all participants for each question in the WESAD dataset while training on a randomly sampled subset of $10$ data points in both SSL and supervised modes. We conduct bootstrapped sampling using $5$ independent samples, and the same samples are used to compare the SSL pre-trained and purely supervised models. From Figure~\ref{15pointsresult}, we observe that in most of the cases, the RMSE score is lower for SSL-based models than purely supervised trained models. We observe that SSL models outperform the purely supervised models in approximately $90\%$ of cases.

\subsection{Results for a demonstrative user}{\label{Results for a demonstrative user}}

\begin{figure*}

  \includegraphics[width=\linewidth]{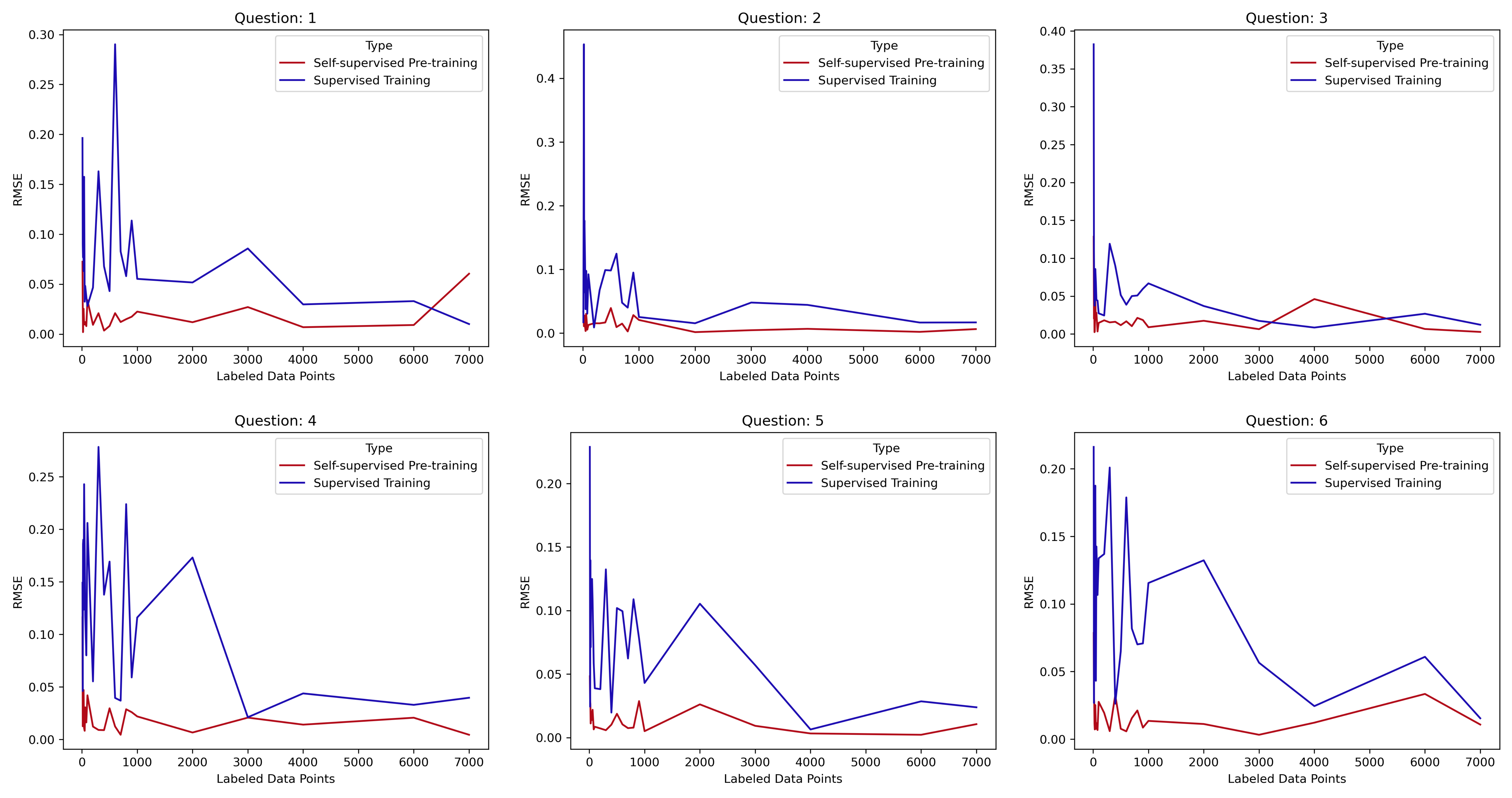}
  \caption{Comparison of the performance between SSL pre-training vs. purely supervised training for Subject 2 for all questions over different sets of labeled data points. We show these results for all participants in the WESAD dataset in our Supplementary Material. }
  \label{error}
\end{figure*}

We compare purely supervised training and SSL as shown in Figure~\ref{error}. Our study shows that models that use SSL techniques require much fewer data to perform as well as models that do not use SSL techniques. We observe that models that employ SSL techniques require less than 30\% of the data needed by models that do not use SSL techniques.

We see that models that only use fully supervised training methods need more data to get to the same RMSE as similar SSL models that are trained with the same number of labeled data points. This illustrates how SSL methods can speed up data-heavy applications by making them more efficient.

In addition to the benefits of preserving data, our findings demonstrate that SSL methods are stable across training runs. The RMSE of the purely supervised method varies drastically (has a high standard deviation) across bootstrapped samples, but the performance of SSL models is more stable. The fluctuation of the purely supervised model indicates that its performance (as measured by RMSE or other metrics) can vary significantly across different training runs or data samples. This variability can be problematic in real-world situations where there may be limited labeled data or data of varying quality. In contrast, SSL methods offer a more stable way of learning, as the performance of SSL models is less susceptible to fluctuations and remains more consistent across training runs or data samples. This stability is particularly useful in situations where data is limited or noisy, as it allows for more efficient and reliable model training. Therefore, the finding that SSL models are more stable than non-SSL models is an important contribution of this study, highlighting the potential benefits of using SSL methods in real-world applications.

\section{Discussion}

\subsection{Implications for Mobile Sensing Systems}

Mobile sensing systems have traditionally been unable to detect psychiatric and mental health outcomes such as stress due to the inherent subjectivity of the labels and variability across users. SSL on biosignals, such as the methodology introduced and evaluated here, provides an opportunity for mobile sensing solutions which detect complex psychiatric events in a personalized manner. Personalization of models is one of the leading challenges and opportunities in machine learning for healthcare and digital psychiatry \cite{washington2022challenges, washington2023review}. From a user interaction standpoint, users are only required to provide a few (on the order of ten) labeled examples of the adverse event of interest. 

Through the use of a limited number of labeled examples for the targeted adverse event, this methodology effectively reduces the user labeling burden while simultaneously facilitating the development of more tailored and precise models. This approach not only improves the overall user experience but also amplifies the probability of user engagement and adherence to digital therapies due to the improved user experience required to achieve a high-performing personalized mobile sensing model.

Personalization through SSL is likely to work in other domains as well. There are numerous digital healthcare research efforts which are attempting to build machine learning models for recurring conditions and events which are traditionally infeasible for machine learning due to the subjectivity of the classes. For example, some digital therapies for pediatric autism use computer vision models to predict facial expressions evoked by conversational partners \cite{washington2017superpowerglass, washington2016wearable, kline2019superpower, voss2019effect, voss2020designing, voss2016superpower, daniels2018exploratory, daniels2018feasibility}. Often, such digital therapeutic systems involve the curation of user-centric data \cite{kalantarian2019guess, penev2021mobile, washington2020training, kalantarian2018gamified, hou2021leveraging, kalantarian2019labeling, kalantarian2018mobile, kalantarian2020performance}, frequently with the end goal of developing a predictive model which analyzes the behavior of the end user \cite{washington2022improved, deveau2022machine, chi2022classifying, washington2020data, banerjee2021training, banerjee2023training}. Such systems provide a clear opportunity for self-supervised personalization as described here. Personalizing the emotion recognition model powering such digital therapies towards frequent conversational partners by the child is possible with the methods described here. 

The interaction paradigm demonstrated here is similar to how Apple's Face ID works. Once the face recognition system is set up with only a few examples provided by the end users, it can be used repeatedly to unlock the device. Our primary goal of this paper is to demonstrate that this user interaction workflow can be adapted to mobile and wearable health settings.

In terms of implementation platforms, the HCI paradigm demonstrated here can be applied to a wide range of systems, such as smartphone apps, wearable devices, and laptops. For example, a smartphone app could use our technique to provide real-time stress monitoring and personalized tips for dealing with stress, while a wearable device could use the model to identify stress levels during physical activities or to monitor stress levels throughout the day.

The choice of platform depends on a number of factors, such as the specific use case, user preferences, and the desired amount of user activity. For example, a smartphone app might be preferred by individuals who want to actively track their stress levels and receive feedback, while a wearable device might be preferred by people who want to passively track their stress levels and have less contact.

\subsection{Limitations and Future Work}

While this work demonstrates promise for personalized learning for mobile sensing, our evaluation involved only a single dataset (WESAD) and a single modality (EDA). In future work, we would like to see our evaluation expanded to numerous biosignal datasets and health outcomes. Moving forward, an important direction for further investigation is to examine the integration of multiple signals within a single model, potentially enhancing its predictive capabilities and overall performance. Additionally, the importance of explainability in the medical informatics domain cannot be overstated. In future research, there is considerable potential to extend this study by exploring how different features influence our personalized model, thus enhancing its interpretability and applicability throughout the field. It is unclear whether the personalized SSL we presented will generalize to other health events which are not as strongly correlated as EDA and stress. If the method does generalize, then this work has the potential to become widely adopted in several HCI fields including wearable computing, affective computing, and mobile sensing.

There are inherent limitations to the WESAD dataset which we used for evaluation. WESAD provides data from controlled experimental situations where realistic emotional engagement is a challenge. As the participants were required to assess their emotional state after each condition (i.e. baseline, amusement, stress, meditation), there is no continuous annotation of the participant's emotional state in this dataset. Therefore, there is a possibility of not obtaining the participant's actual affective state after each task. During data collection, some variables, such as a participant's proficiency in spoken English and familiarity with the provided subject under the stress condition, may introduce bias. Also, the participants' ages ranged from 25 to 30, which may have introduced bias into the findings. 

In future work, we recommend testing the system using real-time, unimodal data obtained from participants in a more comprehensive and unconstrained experimental setup. This approach could potentially lead to a better understanding of the system's performance and generalizability in real-world scenarios, further strengthening the practical applicability of our findings. Nevertheless, WESAD is a frequently used dataset in the field of affective computing using wearable data \cite{10.1007/978-3-030-85613-7_17,gu2021survey,bajpai2020evaluating,rovinska2022affective}.

\section{Conclusion}

We present a novel method for personalizing deep learning models for the prediction of subjective outcomes such as stress using continuous biosignals. We have demonstrated the low labeling requirement for fine-tuning the model after self-supervised pre-training. This method can enable high-performance personalized modeling with minimal manual annotation effort from end users. This technique enables deep learning models to be tailored to an individual's specific physiological reactions, resulting in more precise and personalized stress prediction. This has the potential to significantly increase the efficacy of stress management and preventive digital interventions while also reducing the load of human annotation that is generally necessary for tailored models. Overall, our findings show that self-supervised learning can enable increasingly individualized mobile sensing solutions.

\section{Declarations}

\subsection{Ethics approval and consent to participate}
Not applicable

\subsection{Consent for publication}
Not applicable
\subsection{Availability of data and materials }
The dataset analysed during the current study is available in \href{https://ubicomp.eti.uni-siegen.de/home/datasets/icmi18/}{Dataset Link}
\subsection{Competing interests}
Not applicable
\subsection{Funding}
Not applicable
\subsection{Authors' contributions}
Both authors have equal contributions to this paper.

\subsection{Acknowledgements}

The technical support and advanced computing resources from the University of Hawaii Information Technology Services – Cyberinfrastructure, funded in part by the National Science Foundation CC* awards \# 2201428 and \# 2232862 are gratefully acknowledged.

\subsection{Author's information}
Tanvir Islam is a Masters's student in the Department of Information and Computer Sciences at the University of Hawaii at Manoa. He is primarily focused on conducting research within the healthcare sector, specifically in the areas of Natural Language Processing (NLP) and Bio-signals. He is employed as a student researcher at both the Hawaii Digital Health Lab and the Laboratory for Advanced Visualization \& Applications (LAVA) at the University of Hawaii at Manoa.

Peter Washington is a tenure-track Assistant Professor in the Department of Information and Computer Sciences at the University of Hawaii at Manoa. His work spans digital health and machine learning innovations for healthcare with a particular focus on underserved populations in Hawaii. Prior to starting the Hawaii Digital Health Lab, he completed his PhD in Bioengineering from Stanford University, MS in Computer Science from Stanford University, and BA in Computer Science from Rice University.

\bibliography{sn-article}

\end{document}